\pdfoutput=1

\documentclass[11pt]{article}

\usepackage{EMNLP2022}

\usepackage{times}
\usepackage{latexsym}

\usepackage[T1]{fontenc}

\usepackage[utf8]{inputenc}

\usepackage{microtype}
\usepackage{adjustbox}
\usepackage{amstext}
\usepackage{amsmath}
\usepackage{amssymb}
\usepackage{makecell}
\usepackage{multirow}
\usepackage{multicol}
\usepackage{verbatim}
\usepackage{booktabs}
\usepackage{kotex}
\usepackage[boxed]{algorithm}
\usepackage{algorithmicx}
\usepackage{algpseudocode}
\usepackage[flushleft]{threeparttable}

\usepackage{inconsolata}

\title{Keep Me Updated! Memory Management in Long-term Conversations}

\author{\textbf{Sanghwan Bae$^{1,2}$ \ \ Donghyun Kwak$^{1,2}$ \ \ Soyoung Kang$^{1,2}$ \ \ Min Young Lee$^{1,2}$} \\ \textbf{Sungdong Kim$^{2,3}$ \ \ Yuin Jeong$^1$ \ \ Hyeri Kim$^1$ \ \ Sang-Woo Lee$^{1,2,3}$} \\ \textbf{Woomyoung Park$^{1,2}$ \ \ Nako Sung$^1$} \\
    \\
   NAVER CLOVA$^1$ \ \ NAVER AI Lab$^2$ \ \ KAIST AI$^3$\\
}

\begin{document}
\maketitle
\begin{abstract}

Remembering important information from the past and continuing to talk about it in the present are crucial in long-term conversations. However, previous literature does not deal with cases where the memorized information is outdated, which may cause confusion in later conversations. To address this issue, we present a novel task and a corresponding dataset of memory management in long-term conversations, in which bots keep track of and bring up the latest information about users while conversing through multiple sessions. In order to support more precise and interpretable memory, we represent memory as unstructured text descriptions of key information and propose a new mechanism of memory management that selectively eliminates invalidated or redundant information. Experimental results show that our approach outperforms the baselines that leave the stored memory unchanged in terms of engagingness and humanness, with larger performance gap especially in the later sessions.

\end{abstract}

\section{Introduction}

In human interactions, memory is an important mechanism that helps us hold conversations, develop rapport, and maintain long-term relationships \cite{alea2003you, nelson2003self, brewer2017remember}.
To this end, recent studies \cite{wu-etal-2020-getting, xu-etal-2022-beyond, xu-etal-2022-long} on open-domain dialogues have proposed methods to remember and utilize persona information \cite{zhang-etal-2018-personalizing} of the interlocutors obtained from previous conversations.
Specifically, they summarize the persona information in an extractive or abstractive way and give it as a condition for generating responses in subsequent conversations.
They show that this feature leads to better consistency and engagingness of the chatbot systems.

\begin{figure}[t]
\centering
\includegraphics[width=\columnwidth]{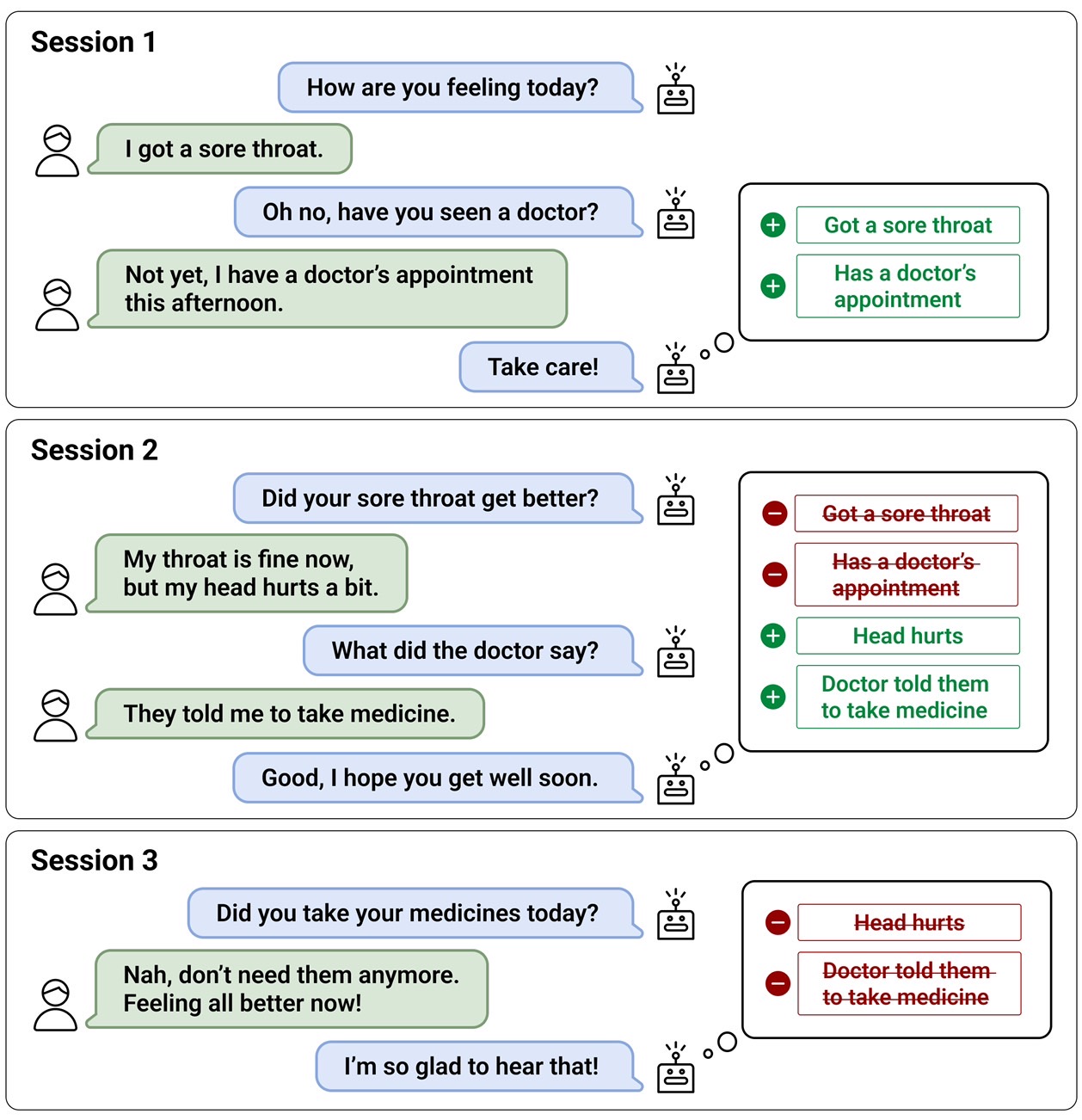}
\caption{An example of a long-term dialogue. There is information obtained from an early session that is no longer true in a later session, e.g. ``Got a sore throat''. This information should be removed from the memory of later sessions in order to correctly follow up with the interlocuter.}
\label{fig:problem}
\end{figure}

Despite such progress, an aspect overlooked by previous studies is that memorized information can be invalidated by newly gathered information.
They simply accumulate and maintain the stored information in memory; once stored, such information has no possibility of getting updated in the future.
Memory in real-life conversations, however, can change over time, either in a short period of time (e.g. health status, plans for the weekend, or recently watched movie) or in relatively longer period of time (e.g. age, job, or hobby). %
Such memory needs to be kept track by asking its status again in subsequent conversations, as exemplified in Figure \ref{fig:problem}.
Therefore, updating previous memory with new relevant information and maintaining it up-to-date are important features of human-like long-term conversations.

In this work, we study the methods of memorizing and updating dynamic information and utilizing them in successive dialogues. 
We formulate a new task of memory management in long-term conversations and construct its corresponding dataset\footnote{The dataset is available at \url{https://github.com/naver-ai/carecall-memory}}, by extending an existing Korean open-domain dialogue dataset \cite{bae-etal-2022-building} to multiple sessions with changing user information.
In each session of our dataset, while the user and the bot have a conversation, information about the user is identified from the dialogue. %
Then, in successive sessions, the bot keeps in memory only the information valid at that point and utilizes the resulting memory in dialogue.

In addition, we propose a long-term dialogue system including a novel memory management mechanism.
In this system, information about the interlocutors revealed in the previous conversation is abstractively summarized and stored in memory.
Specifically, the memory management mechanism decides which information to keep in memory.
For this purpose, we define four pairwise operations (PASS, REPLACE, APPEND, and DELETE) to find and eliminate the information that can cause confusion or redundancy in later conversations.
For example, if the previous memory sentence is ``Haven't got COVID tested yet'' and the new incoming summary is ``Just got positive results from COVID test'', the two sentences are contradictory, in which the former needs to be replaced in memory by the latter.
Through this process, only valid information remains in new memory.
Then, in subsequent sessions, a relevant information from this memory is retrieved and given as additional condition for generating chatbot responses.

With extensive experiments and ablations, we show that the proposed memory management mechanism becomes more advantageous in terms of memorability as the sessions proceed, leading to better engagingness and humanness in multi-session dialogues.

Our contributions are as follows:

\begin{enumerate}
\item We make a step towards long-term conversations with dynamic memory that must be kept up-to-date.
\item We propose a novel memory management mechanism in the form of unstructured text that achieves better results in automatic and human evaluation over baselines.
\item We release the first Korean long-term dialogue dataset for further research on memory management in dialogues.
\end{enumerate}

\section{Related Work}
\paragraph{Personalized Dialogue System}
Building human-like open-domain chatbots is one of the seminal research topics in the field of natural language processing. \citet{zhang-etal-2020-dialogpt} has provided a strong backbone generator model for dialogue systems, while \citet{adiwardana2020towards}, \citet{roller-etal-2021-recipes} and \citet{thoppilan2022lamda} have paved the way for the development of more human-like, natural-sounding chatbots. The applications of open-domain chatbots have also widely expanded, including role-specified \cite{bae-etal-2022-building} and personalized \cite{zhang-etal-2018-personalizing} dialogue systems.
In particular, personalized dialogue system has typically been studied either via utilizing predefined, explicitly stated user profile \cite{zhang-etal-2018-personalizing}, or via directly extracting user profile from dialogue history \cite{xu-etal-2022-beyond, xu-etal-2022-long}.
While the latter approach is preferred in recent research works \cite{zhong-etal-2022-less}, long-term management of the obtained information is yet to be studied.
\paragraph{Long-term Memory in Conversation}
Because it is inefficient to use the entire dialogue history as long-term memory, techniques for obtaining and managing information from dialogue history have been studied. Representing latent features as neural memory \cite{weston2015memory, tran-etal-2016-recurrent, munkhdalai2019metalearned} used to be a traditional method.
Slot-value format in dialogue state tracking \cite{heck-etal-2020-trippy, hosseini2020simple, kim-etal-2020-efficient}, and graph format in \citet{hsiao2020hybrid} have been the two major approaches in handling the memorized information in a structured way. \citet{kim-etal-2020-efficient} suggested update operations on fixed-sized slot-value pairs for dialogue states. \citet{wu-etal-2020-getting} extracted user attributes from dialogues in triples. However, such approaches have not been demonstrated in a multi-session setting.

Leveraging the advancement of pre-trained language models \cite{devlin-etal-2019-bert, raffel2020exploring, NEURIPS2020_1457c0d6, kim-etal-2021-changes}, recent studies attempt to use the unstructured form of text as memory, which is expected to be advantageous in terms of generalizability and interpretability.
\citet{ma2021one} and \citet{xu-etal-2022-long} selectively stored dialogue history with relevant information, while \citet{zhong-etal-2022-less} employed refiners to extract fine-grained information from dialogue history. \citet{xu-etal-2022-beyond} summarized the dialogue history to avoid overflow and redundancy. Nevertheless, these works rarely consider that the obtained information may change and become outdated.
Specifically, MSC \cite{xu-etal-2022-beyond} does not reflect the change of information. In other words, information in MSC remains fixed once it is stored.
DuLeMon \cite{xu-etal-2022-long} is not formatted in a multi-session manner, making it impossible to track memory changes across multiple sessions.

\section{Task and Dataset}
This section describes the task of long-term conversations with dynamic memory changes and the process of constructing a new dataset to conduct research on this task.

\subsection{Task Definition}
An episode consists of multiple consecutive dialogue sessions with a specific user.
Dialogue context of the current session is $D_t = \{c_1, u_1, c_2, u_2, \cdots, c_t, u_t\}$ at time step $t$, where $c$ and $u$ represent the chatbot’s and user's utterance, respectively.
Natural language memory sentences $M = \{m_1, m_2, \cdots, m_n\}$ contain user information abstracted from the previous sessions of the same episode. Then, given the dialogue context $D_t$, and memory $M$, we are interested in predicting the chatbot's response $c_{t+1}$. At the end of each session, the entire session $D$ is summarized into several sentences of user information, denoted as $S = \{s_1, s_2, \cdots, s_k\}$. Memory sentences $M'$ for the next session are constructed by combining $M$ and $S$.

\begin{table}
\centering
\begin{adjustbox}{max width=\columnwidth}
    \begin{threeparttable}
    \begin{tabular}{lr}
    \toprule
    \textbf{Statistics} \\
    \midrule
    Sessions          & 7,665 \\
    \hspace{0.5cm} Session 1 & 2,812 \\
    \hspace{0.5cm} Session 2 & 2,798 \\
    \hspace{0.5cm} Session 3 & 743 \\
    \hspace{0.5cm} Session 4 & 674 \\
    \hspace{0.5cm} Session 5 & 638 \\
    \midrule
    Turns             & 160,191 \\
    Avg. turns per session           & 20.90 \\
    Avg. words per turn         & 4.93 \\
    Unique words for all turns         & 59,434 \\
    Distinct-1/2 for all turns           & 0.0753/0.2891 \\
    \midrule
    Avg. memory sentences per session $|M|$ & 3.41 \\
    Avg. summary sentences per session $|S|$ & 2.88 \\
    Avg. words per summary sentence & 4.70 \\
    Distinct-1/2 for all summary sentences & 0.1425/0.3926 \\
    \bottomrule
    \end{tabular}
    \end{threeparttable}
\end{adjustbox}
\caption{Statistics of our CareCall$_{mem}$ dataset. Distinct-1/2 \cite{li-etal-2016-diversity} is the number of distinct uni- or bi-brams divided by total number of words.}
\label{tab:stat}
\end{table}

\subsection{Dataset Construction}
To study this task, we build a new dataset based on CareCall dataset\footnote{\url{https://github.com/naver-ai/carecall-corpus}} \cite{bae-etal-2022-building}, which consists of single sessions of open-domain dialogues between bots and users. We choose this dataset because the sessions contain various topics that are likely to change in a short period of time, such as user's health, sleep, and diet, as well as those in a relatively longer period of time, such as family, pets, and frequently visited places. We extend this single-session dataset to a multi-session setting, which is a similar procedure presented in MSC \cite{xu-etal-2022-beyond}. Our resulting dataset contains more persona updates than other datasets \cite{xu-etal-2022-beyond, xu-etal-2022-long} (see Section \ref{sec:discuss_dataset} in Appendix for more details).

\subsubsection{Preliminary Step: Dialogue and Summary}
\label{sec:data_prep}
To efficiently collect the dataset, we train preliminary models for dialogue summaries and memory grounded dialogues to first automatically generate the dataset, and then a group of  annotators revise them.
This procedure has shown to be more effective in recent studies \cite{sun-etal-2021-adding, bae-etal-2022-building, liu2022wanli, zheng2022augesc}.
In the entire process, we leverage the large-scale language models (LMs) for each step; HyperCLOVA 6.9B as backbone LM.

\paragraph{Dialogue Summary}
We randomly sample 600 dialogue sessions with more than 15 turns from the CareCall dataset. We ask annotators to summarize each session into several sentences to build $S$ that may be useful to continue the next conversation.
Using these summaries, we fine-tune LMs to generate summaries given dialogues $P(S|D)$.
The models then generate summaries of unseen dialogues randomly sampled from the CareCall dataset. Finally, annotators edit the generated summaries by filling in missing information or correcting erroneous sentences. Since there is no memory sentence for the first session, i.e. $M = \varnothing$, memory for the second session $M'$ is equal to $S$.

\begin{figure*}[t]
\centering
\includegraphics[width=0.95\textwidth]{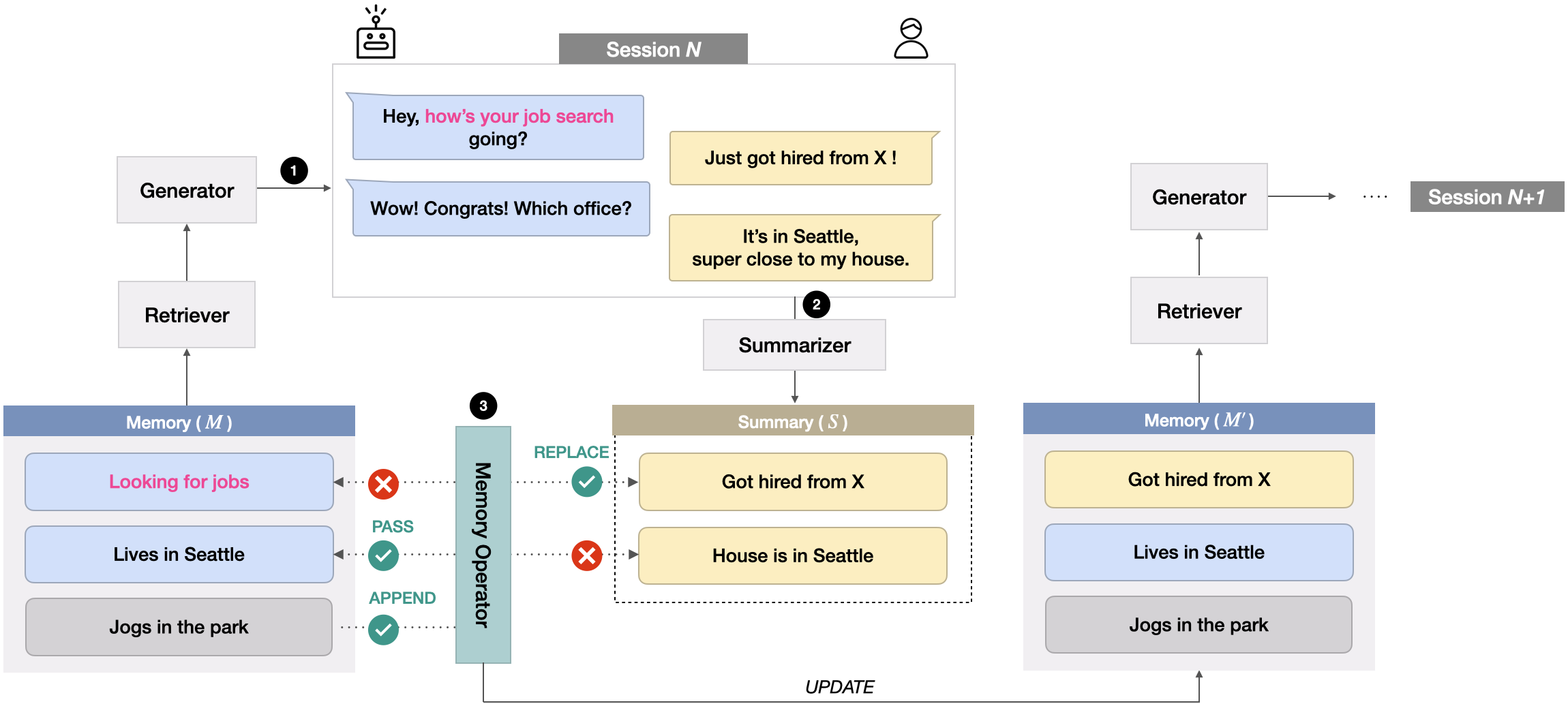}
\caption{The overview of the proposed system. (1) Memory grounded response generation model (Section \ref{sec:gen}) conditioned on memory sentences $M$ converses with human user. (2) At the end of the session, the dialogue summarizer (Section \ref{sec:summ}) summarizes user information into several sentences $S$ from the session history. (3) Memory operator (Section \ref{sec:update}) predicts the operations for every $(m_i, s_j)$ pair to select information to leave, which consists the next memory $M'$.}
\label{fig:pipeline}
\end{figure*}

\paragraph{Memory Grounded Dialogue}
To build a second session of each episode, annotators write dialogue sessions grounded on the 600 human-written summaries from the previous step. Likewise, we fine-tune LMs to generate the entire dialogue sessions given previous memory $P(D|M)$. Then, the fine-tuned models generate memory grounded dialogues from the unseen dialogue summaries in the previous paragraph. Lastly, human annotators revise the generated dialogues, i.e. correcting wrong responses (misuse of memory, not sensible, or out-of-bounds from CareCall's role described in \citet{bae-etal-2022-building}).

\subsubsection{Interactive Step: Multi-Session Dialogue}
\label{sec:data_multi}
From the preliminary step, we obtain the data to build a chatbot that can conduct interactive conversation utilizing the memorized information. To construct a multi-session dialogue system, we train the dialogue summarizer and memory grounded response generator described in Section \ref{sec:model} on previously collected $(D, S)$ pairs with $(M, D)$ pairs respectively.

Then, crowdworkers converse with the resulting system for 5 sessions per episode, starting from the first session.
The interval between sessions is assumed to be from 1 to 2 weeks.
At the end of each session, the summarizer generates $S$ from the current session.  
Both generated responses and summaries are edited by annotators to correct errors.
Lastly, we ask annotators to select which sentences in $M$ and $S$ should remain in new memory $M'$ for the next session.
We provide details of quality control in Appendix \ref{sec:quality} and an example episode in Figure \ref{fig:example} in Appendix.
We name this dataset as CareCall$_{mem}$ and the statistics of the dataset are given in Table \ref{tab:stat}, which includes all the collected data described in Section \ref{sec:data_prep}-\ref{sec:data_multi}.

\section{Models}
\label{sec:model}
We propose a long-term dialogue system with memory management mechanism.
The system consists of three parts: memory grounded response generation, dialogue summarization, and memory update.
The overall architecture is shown in Figure \ref{fig:pipeline}.

\subsection{Memory Grounded Response Generation}
\label{sec:gen}

\paragraph{Response Generation}
We consider the response generation model conditioned on memory sentences.
Given the memory $M$ and the dialogue history $D_t = \{c_1, u_1, c_2, u_2, \cdots, c_t, u_t\}$ at time step $t$, the conditional probability of the next target response $c_{t+1} = \{w_1, w_2, \cdots, w_{|c_{t+1}|}\}$ can be written as the product of a sequence of conditional probabilities:
\begin{equation}
p(c_{t+1}|D_t, M) = \prod_{i} p_\theta(w_i|D_t, M, w_{<i}),
\end{equation}
where $w_i$ is $i$-th token of the sequence and $\theta$ is trainable parameters of the model.
We use HyperCLOVA 6.9B as the response generation model. The model is fine-tuned using the maximum likelihood estimation (MLE), which minimizes:
\begin{equation}
\mathcal{L}_\theta(c_{t+1}, D_t, M) = - \sum_{i}{\log{p_\theta(w_i|D_t, M, w_{<i})}}.
\end{equation}

\paragraph{Memory Retrieval}
In addition, following the previous studies \cite{xu-etal-2022-beyond, xu-etal-2022-long}, we consider that retrieving information relevant to the current dialogue context is effective when dealing with a large collection of sentences in memory.
We use an approach almost identical to context persona matching (CPM) method proposed in \citet{xu-etal-2022-long}, replacing persona sentences to memory sentences in our task (See Appendix \ref{sec:detail_ret} for more details).
At the inference time, the retrieved top $k$ sentences constitute $M_{ret}$, which is the actual input condition of the response generator described in the preceding paragraph.

\subsection{Dialogue Summarization}
\label{sec:summ}
Given the dialogue history of the entire session $D$, our abstractive summarization model summarizes important user information in the form of several natural language sentences $S = \{s_1, s_2, \cdots, s_k\}$.
What information to be summarized can be learned based on the human annotation from our newly collected CareCall$_{mem}$ dataset.
Here, we train HyperCLOVA 6.9B as the summarizer to generate summary sentences given the dialogue history as an input.
Formally, this is done by minimizing loss for each gold summary sentence $s_t = \{w_1, w_2, \cdots, w_{|s_t|}\}$:
\begin{equation}
\mathcal{L}_\phi(s_t, D, s_{<t}) = - \sum_{i}{\log{p_\phi(w_i|D, s_{<t}, w_{<i})}},
\end{equation}
where $\phi$ is trainable parameters of the summarizer.

\begin{algorithm}[t!]
\textbf{Input} $M =\{m_1, m_2, \cdots, m_n\}$ \\
\textbf{Input} $S =\{s_1, s_2, \cdots, s_k\}$ \\
\textbf{Input} $O(m, s) \rightarrow \{``\text{P}", ``\text{R}", ``\text{A}", ``\text{D}"$\}\\
\textbf{Output} $M' = \{m'_1, m'_2, ..., m'_{|M'|}\}$
\begin{algorithmic}[1]
\State $M_{del} \leftarrow \varnothing$
\State $S_{del} \leftarrow \varnothing$
\State \textbf{for} $\forall m_i \in M$:
\State \hspace{0.5cm} \textbf{for} $\forall s_j \in S$:
\State \hspace{1cm} \textbf{if} $O(m_i, s_j) \in \{``\text{R}", ``\text{D}"\}$:
\State \hspace{1.5cm} $M_{del} \leftarrow M_{del} \cup \{m_i\}$
\State \hspace{1.5cm} \textbf{if} $O(m_i, s_j) = ``\text{D}"$:
\State \hspace{2.0cm} $S_{del} \leftarrow S_{del} \cup \{s_j\}$
\State $M \leftarrow M - M_{del}$
\State \textbf{for} $\forall s_j \in S$:
\State \hspace{0.5cm} \textbf{for} $\forall m_i \in M$:
\State \hspace{1cm} \textbf{if} $O(m_i, s_j) = ``\text{P}"$:
\State \hspace{1.5cm} $S_{del} \leftarrow S_{del} \cup \{s_j\}$
\State $S \leftarrow S - S_{del}$
\State $M' \leftarrow M \cup S$ \\
\Return $M'$
\end{algorithmic}
\caption{Our memory update algorithm.
P, R, A, and D are abbreviations for PASS, REPLACE, APPEND, and DELETE, respectively.
}\label{alg:update}
\end{algorithm}

\subsection{Memory Update}
\label{sec:update}
The memory update process stores the latest user information in memory by combining the old and the new information sentences.
At the end of each session, $n$ existing memory sentences $M = \{m_1, m_2, ..., m_n\}$ and $k$ new summary sentences $S = \{s_1, s_2, ..., s_k\}$ are given.
The memory writer combines them to find $M' = \{m'_1, m'_2, ..., m'_{|M'|}\}$ that are lossless, consistent, and not redundant in terms of information.
Here, we assume that $M$ and $S$ are internally consistent and not redundant.

Our approach finds the sentence set $M'$ by classifying the relationship of the sentence pair $(m_i, s_j)$, where $m_i \in M$ and $s_j \in S$.
We define operations for  $(m_i, s_j)$ as $O(m_i, s_j) \rightarrow$ {\footnotesize $\{``\text{PASS}", ``\text{REPLACE}", ``\text{APPEND}", ``\text{DELETE}"\}$}.

\begin{itemize}
  \item \textbf{PASS} %
  means storing only $m_i$. It reflects the case in which the information of $m_i$ already contains that of $s_j$, i.e. $m_i \supseteq s_j$ in terms of information. Only $m_i$ is stored in order to avoid redundancy.
  \item \textbf{REPLACE} means storing only $s_j$. When $s_j$ and $m_i$ are inconsistent, the more recent information $s_j$ remains. This operation is also useful when $s_j$ has more information than $m_i$, i.e. $m_i \subset s_j$ in terms of information.
  \item \textbf{APPEND} means storing both $m_i$ and $s_j$. If $m_i$ and $s_j$ are irrelevant, both are stored in order to avoid loss of information.
  \item \textbf{DELETE} means removing both $m_i$ and $s_j$. We found that there are cases where this operation is useful.
It can reduce the memory confusion by “forgetting” a completed state that no longer needs to be remembered.
For example, if $m_i =$ “having a cold and taking medicine” and $s_j =$ “cold is all better now”, not only $m_i$ should be removed because it is no longer true, but $s_j$ should also be removed because the chatbot doesn't have to remember the user's cold anymore.
If such information is not forgotten and remained in memory, something like Pink Elephant Paradox \cite{wegner1987paradoxical} can occur, causing hallucination of the dialogue model.
Therefore, we decided to delete the information that no longer needs to be remembered.
\end{itemize}

A formal description of the entire algorithm for memory update is given in Algorithm \ref{alg:update}, which use the proposed pairwise operations.
We fine-tune a classification model of T5 \cite{raffel2020exploring} architecture to perform $O(m,s)$.

\section{Experiments}

In this section we describe experimental settings and results including the evaluations of the proposed memory update methods of memory management (Section  \ref{sec:eval_update}) and the evaluations of the entire system on multi-session dialogues (Section \ref{sec:eval_multisession}).

\subsection{Memory Update}
\label{sec:eval_update}

In this section, we compare several types of models to perform $O(m_i, s_j)$ described in Section \ref{sec:update}.

\subsubsection{Pairwise Evaluation}
\label{sec:pairwise}

\begin{table}
\centering
\begin{adjustbox}{max width=\columnwidth}
\begin{tabular}{lr}
\toprule
\textbf{Statistics} \\
\midrule
Training pairs & 2,149 \\
Validation pairs & 300 \\
Test pairs & 300 \\
\midrule
individual label = gold label & 87.52\% \\
(estimated human performance) \\
no gold label & 1.96\% \\
\midrule
PASS & 13.3\% \\
REPLACE & 37.0\% \\
APPEND & 44.6\% \\
DELETE & 5.1\% \\
\bottomrule
\end{tabular}
\end{adjustbox}
\caption{Statistics of the collected sentence pairs to train and evaluate memory update methods.}
\label{tab:pairwise_statistics}
\end{table}

To build a dataset to train and evaluate, we ask the annotators to annotate each pair of $(m_i, s_j)$ in CareCall$_{mem}$ dataset into 4 classes {\footnotesize $\{``\text{PASS}", ``\text{REPLACE}", ``\text{APPEND}", ``\text{DELETE}"\}$}.
If any one of the four labels is chosen by at least two of the three annotators, it is regarded as the gold label.
If there is no such consensus, which occur in about 2\% of the cases, we discard the example.
Table \ref{tab:pairwise_statistics} reports some key statistics about the collected dataset.
Examples of the annotated pairs are in Table \ref{tab:pairwise_example} of the Appendix and see Appendix \ref{sec:discuss_operation} for discussions of the cases we found that need further research on this direction.

We consider three types of model for this task.
A pre-trained T5 architecture is used for all three models (See Appendix \ref{sec:detail_operator} for more details). 

\begin{itemize}
  \item \textit{From scratch}: A model fine-tuned on the collected dataset.
  \item \textit{NLI zero-shot}: A model fine-tuned on KLUE-NLI \cite{park2021klue}, a Korean natural language inference (NLI) dataset. 
Assuming the old memory sentence and the new memory sentence as a premise and a hypothesis, respectively, we can map the memory update operations as follows.{\footnotesize $``\text{PASS}"$}: entailment, {\footnotesize $``\text{REPLACE}"$}: contradiction or reversely entailment, {\footnotesize $``\text{APPEND}"$}: neutral. {\footnotesize $``\text{DELETE}"$} is not mapped, so this model has minimum 5.1\% of error. The \textit{NLI zero-shot} is used to see if the knowledge from NLI can be transferred to the proposed memory operations.
  \item \textit{NLI transfer (fine-tune)}: The \textit{NLI zero-shot} model further fine-tuned on the collected dataset.
\end{itemize}

\begin{table}
\centering
\begin{adjustbox}{max width=\columnwidth}
\begin{tabular}{lccc}
\toprule
& \multicolumn{2}{c}{\textbf{Pairwise Acc.}} & \textbf{Set F1} \\
\cmidrule(lr){2-3}\cmidrule(lr){4-4}
\textbf{Model} & \textbf{Validation} & \textbf{Test} & \textbf{Test} \\
\midrule
From scratch & 84.65 (0.99) & 83.65 (2.01) & 87.98 (2.11) \\
NLI zero-shot & 72.23 (1.61) & 71.50 (1.24) & 84.62 (2.20) \\
NLI transfer (fine-tune) & \textbf{85.34} (0.81) & \textbf{84.10} (1.01) & \textbf{88.69} (1.65)\\
\bottomrule
\end{tabular}
\end{adjustbox}
\caption{4-class accuracy for pairwise operations and sentence-level F1 scores for set-level evaluation (standard deviation in brackets).}
\label{tab:pairwise}
\end{table}

The results are shown in Table \ref{tab:pairwise}.
Although \textit{NLI zero-shot} reaches certain level of performance, there is a significant performance drop compared to \textit{From scrach} model ($-$12.15\% on test set).
We hypothesize that this is because the memory update requires common sense beyond pure logical reasoning.
For example, if $m_i =$ ``planning to see a doctor'' and $s_j$ = ``went to the hospital'', there is no logical inconsistency, but $m_i$ can generally be expected to be replaced by $s_j$.
Eventually, we found that further training the NLI model on the collected dataset is the best (\textit{NLI transfer}).

\begin{table*}[hbt!]
\centering
\begin{adjustbox}{max width=\textwidth}
\begin{tabular}{lrrrrrrrrrrrr}
\toprule
\textit{All turns} & \multicolumn{3}{c}{\textbf{Session 2}} & \multicolumn{3}{c}{\textbf{Session 3}} & \multicolumn{3}{c}{\textbf{Session 4}} & \multicolumn{3}{c}{\textbf{Session 5}} \\
\cmidrule(lr){2-4}\cmidrule(lr){5-7}\cmidrule(lr){8-10}\cmidrule(lr){11-13}
\textbf{Model} & \textbf{PPL$\downarrow$} & \textbf{BLEU-1/2$\uparrow$} & \textbf{F1$\uparrow$} & \textbf{PPL$\downarrow$} & \textbf{BLEU-1/2$\uparrow$} & \textbf{F1$\uparrow$} & \textbf{PPL$\downarrow$} & \textbf{BLEU-1/2$\uparrow$} & \textbf{F1$\uparrow$} & \textbf{PPL$\downarrow$} & \textbf{BLEU-1/2$\uparrow$} & \textbf{F1$\uparrow$} \\
\midrule
Without memory & 4.023 & 0.293/0.169 & 0.334 & 4.789 & 0.298/0.174 & 0.332 & 4.073 & 0.289/0.167 & 0.320 & 4.221 & 0.289/0.159 & 0.334 \\ 
History accumulate & 4.057 & 0.267/0.161 & 0.320 & 4.491 & 0.263/0.153 & 0.322 & 4.652 & 0.261/0.151 & 0.321 & 4.673 & 0.261/0.154 & 0.329 \\
Memory accumulate & \textbf{3.735} & \textbf{0.313}/\textbf{0.189} & \textbf{0.365} & 3.782 & 0.314/0.189 & 0.368 & 3.875 & 0.307/0.186 & 0.358 & 4.052 & 0.311/0.193 & 0.364\\
Memory update & 3.743 & 0.312/0.187 & 0.363 & \textbf{3.773} & \textbf{0.316}/\textbf{0.192} & \textbf{0.369} & \textbf{3.794} & \textbf{0.309}/\textbf{0.188} & \textbf{0.360} & \textbf{3.937} & \textbf{0.316}/\textbf{0.198} & \textbf{0.369}\\
\midrule
Memory gold* & 3.680 & 0.317/0.201 & 0.375 & 3.736 & 0.325/0.206 & 0.383 & 3.746 & 0.318/0.201 & 0.377 & 3.878 & 0.320/0.202 & 0.375 \\
\bottomrule
\toprule
\textit{Memory turns} & \multicolumn{3}{c}{\textbf{Session 2}} & \multicolumn{3}{c}{\textbf{Session 3}} & \multicolumn{3}{c}{\textbf{Session 4}} & \multicolumn{3}{c}{\textbf{Session 5}} \\
\cmidrule(lr){2-4}\cmidrule(lr){5-7}\cmidrule(lr){8-10}\cmidrule(lr){11-13}
\textbf{Model} & \textbf{PPL$\downarrow$} & \textbf{BLEU-1/2$\uparrow$} & \textbf{F1$\uparrow$} & \textbf{PPL$\downarrow$} & \textbf{BLEU-1/2$\uparrow$} & \textbf{F1$\uparrow$} & \textbf{PPL$\downarrow$} & \textbf{BLEU-1/2$\uparrow$} & \textbf{F1$\uparrow$} & \textbf{PPL$\downarrow$} & \textbf{BLEU-1/2$\uparrow$} & \textbf{F1$\uparrow$} \\
\midrule
Without memory & 6.655 & 0.285/0.128 & 0.361 & 6.779 & 0.275/0.127 & 0.358 & 6.577 & 0.279/0.120 & 0.346 & 7.106 & 0.269/0.117 & 0.330 \\
History accumulate & \textbf{4.246} & 0.242/0.131 & 0.339 & \textbf{4.548} & 0.245/0.135 & 0.342 & 5.008 & 0.228/0.126 & 0.323 & 6.617 & 0.219/0.102 & 0.275 \\
Memory accumulate & 4.439 & 0.324/0.160 & 0.381 & 4.620 & 0.304/0.136 & \textbf{0.362} & 5.117 & 0.293/0.125 & 0.354 & 5.725 & \textbf{0.284}/0.117 & 0.333\\
Memory update & 4.487 & \textbf{0.329}/\textbf{0.163} & \textbf{0.382} & 4.627 & \textbf{0.306}/\textbf{0.145} & 0.360 & \textbf{4.872} & \textbf{0.297}/\textbf{0.130} & \textbf{0.360} & \textbf{5.308} & \textbf{0.284}/\textbf{0.120} & \textbf{0.339}\\
\midrule
Memory gold* & 4.419 & 0.352/0.188 & 0.395 & 4.421 & 0.338/0.182 & 0.393 & 4.518 & 0.335/0.178 & 0.386 & 4.855 & 0.322/0.177 & 0.379\\
\bottomrule
\end{tabular}
\end{adjustbox}
\caption{Comparison of automatic evaluation metric results among different systems on test set in multi-session dialogues. \textit{Memory turns} refer to the turns annotated to use memory explicitly, which accounts for 22.4\% of all turns in the test set.}
\label{tab:effectiveness}
\end{table*}

\subsubsection{Set Evaluation}

We manually annotate the test set in Table \ref{tab:pairwise_statistics} to measure the performance of set level algorithm $f(M, S) \rightarrow M'$ (Alg. \ref{alg:update}).
The annotation process is the same as the one described in Section \ref{sec:data_multi}, and the compared models are the same as the pairwise evaluation.
We measure the sentence-level F1 scores between gold $M'$ and the predicted one.
The evaluation results in Table \ref{tab:pairwise} show a trend similar to the pairwise evaluation; \textit{NLI transfer} achieves the best performance.

\subsection{Multi-session Dialogues}
\label{sec:eval_multisession}
We evaluate our entire system described in Section \ref{sec:model} in a multi-session dialogue setting.
The response generation model (\ref{sec:gen}) is trained on both CareCall$_{mem}$ and original CareCall datasets (details are in Appendix \ref{sec:eval_gen}).
The memory retrieval (\ref{sec:gen}) and dialogue summarization (\ref{sec:summ}) models are trained on CareCall$_{mem}$ only.
The memory update model (\ref{sec:update}) is \textit{NLI transfer (fine-tune)} from \ref{sec:eval_update}.

\begin{table*}
\centering
\begin{adjustbox}{max width=0.8\textwidth}
\begin{tabular}{l|rrrrr}
\toprule
\textbf{Model} & \textbf{Coherence} & \textbf{Consistency} & \textbf{Engagingness} & \textbf{Humanness} & \textbf{Memorability} \\
\midrule
Without memory & $-$0.0450 & 0.5907 & $-$0.4625 & $-$0.2445 & $-$1.3057 \\
Memory accumulate & 0.1892 & \textbf{0.6301} & $-$0.2871 & 0.0831 & 0.0030 \\
Memory update & \textbf{0.2248} & 0.6272 & \textbf{$-$0.1770} & \textbf{0.1917} & \textbf{0.4351} \\
\bottomrule
\end{tabular}
\end{adjustbox}
\caption{Average standardized scores of human evaluation metrics on live conversation for 5 sessions in each episode. Absolute scores are in Table \ref{tab:human_abs} in Appendix.}
\label{tab:human}
\end{table*}

\begin{figure*}[t]
\centering
\includegraphics[width=\textwidth]{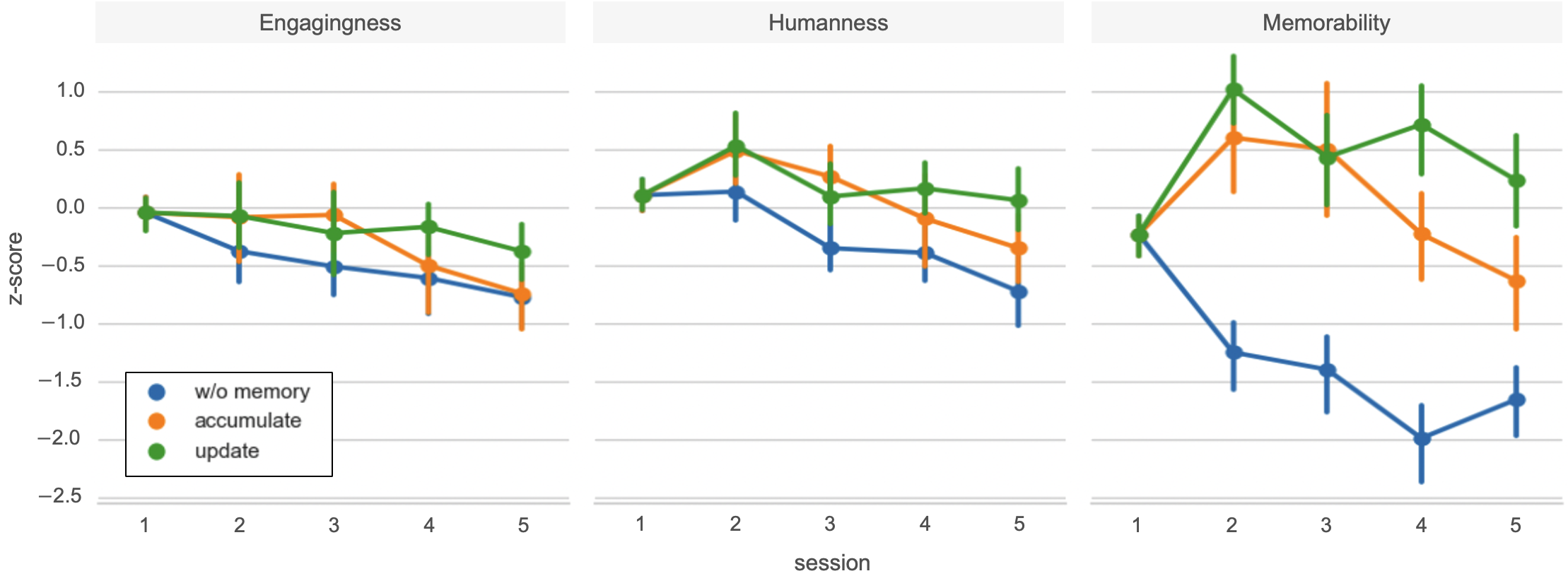}
\caption{Average standardized scores per session. The dots indicate the average scores and the vertical lines indicate the 95\% confidence interval. The averaged scores outside the vertical line of other models mean significant differences at the 95\% confidence level. The models in the first session are all equivalent.}
\label{fig:human}
\end{figure*}

\subsubsection{Automatic Evaluation}
\label{sec:auto}
We randomly sample 60 episodes (300 sessions) from CareCall$_{mem}$ dataset to build a test set.
As evaluation metrics, PPL, BLEU-1/2 \cite{papineni-etal-2002-bleu}, F1, and Distinct-1/2 \cite{li-etal-2016-diversity} are used.
We compare four models in this section.

\begin{itemize}
\item \textit{Without memory}: This model consists of only the response generation model in Section \ref{sec:model}. The input at inference is current dialogue history $D_t$ only.
\item \textit{History accumulate}: This model consists of only a response generation model, with input of all previous sessions concatenated before $D_t$ for both training and inference.
\item \textit{Memory accumulate}: This model consists of all components in Section \ref{sec:model}, but it use {\footnotesize $``\text{APPEND}"$} as the only pairwise operation.
\item \textit{Memory update}: This model consists of all components in Section \ref{sec:model}. The entire memory update method in Algorithm \ref{alg:update} is used.
\item \textit{Memory gold*}: This model consists of all components in Section \ref{sec:model}, but $M$ for each session is gold memory in the dataset. This serves as the upper limit for memory management mechanism.
\end{itemize}

Table \ref{tab:effectiveness} shows the results.
First, \textit{Without memory} shows a relatively high PPL in memory turns of all sessions.
\textit{History accumulate} is competitive in all metrics in early sessions, but its performance drops significantly as the session progresses. %
This is conjectured that there are many distractions in accumulated dialogue history which make it difficult to track changing information.
Furthermore, the performance gain of \textit{Memory update} over \textit{Memory accumulate} becomes larger as the sessions progress.
This is because \textit{Memory accumulate} has a relatively high possibility of utilizing outdated information, which can be a noise to the model.
\textit{Memory update} achieves the best performance in most metrics especially in later sessions (Session 4-5), showing advantage of up-to-date memory.

\subsubsection{Human Evaluation}

We also perform human evaluation in multi-session dialogues.
For a reliable evaluation, we use continuous rating for the live conversations proposed in \citet{ji-etal-2022-achieving}, which is shown to be easily reproducible with high correlation among repeated experiments.
We extend this evaluation process to multi-session dialogues.
A crowdworker conducts live conversations for five sessions per episode (assuming 1-2 weeks are elapsed between sessions) with a randomly selected model among the compared models.
At the end of each session, we ask the crowdworkers to rate the degree to which they agree with the statements on each evaluation metric on a scale of 0-100.
After an episode is over, they repeat another episode with another randomly selected model.
The score distribution of each crowdworker is standardized, removing the potential bias of each worker.
The evaluation metrics are summarized in Appendix \ref{sec:metric} and the interface used for evaluation is shown in Figure \ref{fig:webui} in Appendix.
There are three models compared in this evaluation: \textit{Without memory}, \textit{Memory accumulate}, and \textit{Memory update} described in Section \ref{sec:auto}.
A total of 155 episodes and 775 sessions are evaluated trough this process.

Overall results are shown in Table \ref{tab:human}.
\textit{Memory update} shows a clear advantage over \textit{Memory accumulate} in memorability (p-value < 0.05 for pairwise significance test).
We also discover that memorability positively correlates with engagingness (Pearson correlation 0.68 at p-value < 0.01).
Furthermore, there is some positive correlation between memorability and humanness (Pearson correlation 0.47 at p-value < 0.01).
Accordingly, \textit{Memory update} shows the highest engagingness and humanness compared to the other two models.
Additionally, coherence and consistency have no statistically meaningful difference among the models (p-value > 0.1 for pairwise significance test). This allows us to conclude that \textit{Memory update} has better ability to remember while still preserving general conversational abilities like coherence and consistency. 

Figure \ref{fig:human} shows scores of each session for the three metrics (results for all five metrics are in Table \ref{tab:human_all} in Appendix).
In the case of \textit{Without memory}, it is observed that engagingness and humanness consistently drop as the sessions progress.
In contrast, \textit{Memory accumulate} and \textit{Memory update} maintain engagingness and humanness to some level in the subsequent sessions.
However, in the later sessions (Session 4-5), the difference between \textit{Memory accumulate} and \textit{Memory update} grows with regards to all three metrics.
It seems that the crowdworkers feel as if the chatbot doesn't remember well when it brings up an information that has become no longer true in previous sessions, resulting in a lower engagingness and humanness.
This is likely the reason why crowdworkers rate \textit{Memory update} as the most engaging and human-like.

\section{Conclusion}
We present a novel task of long-term conversation with dynamic memory changes and build the corresponding dataset.
We propose a memory management method that performs operations between old and new memory information in the form of unstructured text.
Through an extensive series of experiments, we demonstrate the effectiveness of the proposed method in terms of improving memorability of a chatbot system.
We also show that keeping memory up-to-date in long-term conversations is important for engaging and human-like dialogues.
We release the newly collected dataset, looking forward to further research on this promising direction.

\section*{Limitations}

For the sake of simplicity and clarity in our current research study, we only considered remembering and updating information of a single interlocutor. However, our future studies should aim to include memorized information from both sides and bringing it up in conversations, just as \citet{xu-etal-2022-beyond, xu-etal-2022-long} did by duplicating the proposed memory management for both sides.

Regarding the generalizablity of our results, it should be noted that the experiment was performed on data collected in Korean language. Although we do not use a Korean-language-specific approach in our experimental settings, whether or not our results would extend across different languages is yet to be determined.

Our experiments do not cover extremely long conversations where memory reaches its maximum capacity.
In this case, removing the oldest memories (i.e. first in, first out) could be a plausible approach, just as human memory fades over time.
Still, the amount of computation can grow large.
Since pairwise operation occurs $\mathbf{O}(|M||S|)$ times and each operation is predicted by T5 in our experiments (it takes about 80ms on 1 NVIDIA V100 for a single inference), the memory update can be costly when $|M|$ gets large.

Finally, our experiments require large GPU resources, at least 1 NVIDIA A100 or multiple GPUs equivalent to it. The specifications of GPUs used for training the models are provided in Appendix \ref{sec:detail}.

\section*{Ethical Considerations}
Our dataset is created by authors, crowdworkers, and large-scale language models.
Throughout the interactive data collection process, we instructed crowdworkers to play the role of potential users only, without disclosing any personally identifiable information about workers themselves.
Meanwhile, it is known that the generated output from pre-trained language models may contain toxicity \cite{gehman-etal-2020-realtoxicityprompts, liu-etal-2021-dexperts, xu-etal-2021-bot}, private information \cite{carlini2021extracting}, or social biases \cite{bordia-bowman-2019-identifying, shwartz-choi-2020-neural, bender2021dangers, garrido2021survey}.
To address these issues, we carefully constructed criteria for harmful texts based on legal and ethical considerations offered by our group's specialists.
We guided all annotators to filter and edit the dataset based on such criteria.
In addition, since the users in our dataset might be deemed as a vulnerable social group, our group's ethical consultation included a review of sensitive subjects and the elimination of sessions involving any mention of such topics.
We also had multiple filtering processes by multiple workers for every example to ensure that the final dataset does not contain any potentially malicious or unintended harmful effects.

Furthermore, since the proposed system has the capability of storing the information they learned from the interactions with users, we emphasize that the information in long-term memory remains absolutely private to the individual's conversation and is not shared with anyone else.
Also, if it comes into the wrong hands, chatbots are exposed to the possibility of getting programmed to imitate humans and be used for phishing and fraud.
In such conversations, we can expect abusive cases where individuals accidentally disclose important and confidential information.
Thus, incorporating the proposed system into real-world applications requires developers to ensure that it is used only in a safe and ethical manner.

\section*{Acknowledgements}
The authors thank all the members of CLOVA and AI Lab of NAVER for devoted support and discussion.
In particular, they would like to thank the members of CLOVA Conversation for their technical support and active discussion.
In addition, the authors thank the members of CLOVA Conversation Planning for guiding and monitoring the data collection process.
Finally, they would like to show appreciation for the valuable feedback given by Professor Yejin Choi.

\bibliography{anthology,custom}
\bibliographystyle{acl_natbib}

\clearpage
\appendix

\section{Data Quality Control}
\label{sec:quality}
In each annotation process, we provide detailed guidance and training for all annotators in order to optimize our dataset qulity. Specifically, we instruct them with a comprehensive manual for each job and clarify all questions in a group chat to eliminate potential misunderstandings.
We also give personalized feedback to each annotator every week.
We recruited annotators from a freelancing platform and in-house labeling services.
The major instructions for the data collection process are summarized as follows.

\paragraph{Dialogue Summary} We ask annotators to summarize the core information of the user in a given dialogue session, particularly the information that is worth continuing with the subsequent sessions.
The resulting summaries are abstractive summaries, as we instruct them not to copy the utterance itself from the dialogues, but write it into a new abstractive sentence.
For the summarized information, we guide the annotators to exclude information about one-off events or overly-detailed information about the user such as ``had three eggs for breakfast'' or ``went to the park at 9:10AM'', given that they are difficult or irrelevant to use in subsequent dialogues.
However, if the user’s information has been changed, such information would be included in the summary (e.g. ``just got married'' or ``recovered from flu'') to keep track of. Three different groups of annotators consecutively edited the summaries, only 1.4\% of which were edited in the last iteration.

\paragraph{Memory Grounded Dialogue}
We ask annotators to make sure that the dialogues contain various daily topics and that the information in memory is utilized in a context as naturally as possible, rather than obsessively mentioning memorized information. In addition, for consistency, we ask them to correct cases where the bot generates responses or questions that may contradict information already included in the memory.
For example, if the bot already knows that the user is hospitalized for back surgery, the bot wouldn’t ask questions such as ``Are there any health issues?''.
Also, the bots are not allowed to mention things that are not in memory as if they do remember those things.
For other remaining instructions, we refer to the specifications in \citet{bae-etal-2022-building} for the consistent role of the bot.

\paragraph{Memory Update}
At the end of each session, annotators are asked to select the sentences in M and S that could be used in the subsequent sessions.
In other words, the statements that are no longer valid or redundant are removed, and the statements that do not conflict with others remain.
Additionally, every time when $m_i$ and $s_j$ are exactly the same in terms of information, even if it might be okay to leave either of them, we guide to leave $m_i$ for consistency of the dataset.

\section{Inplementation Details}
\label{sec:detail}

\subsection{Pre-trained Language Models}

We use three types of Transformer-based pre-trained language models in our experiments.
For response generation (Section \ref{sec:gen}) and dialogue summarization (Section \ref{sec:summ}), we use HyperCLOVA \cite{kim-etal-2021-changes} with 6.9B parameters. The model specification follows \citet{kim-etal-2021-changes} and the implementation is based on Megatron-LM \cite{shoeybi2019megatron}.
For the memory update model (Section \ref{sec:update}), we use a model of T5 \cite{raffel2020exploring} architecture pre-trained on the corpus identical to that of \citet{kim-etal-2021-changes}. 
This model consists of 24 layers, 1024-dimensional embeddings, and 16 attention heads, resulting in total of 822M parameters.
Lastly, for retriever (Section \ref{sec:gen}), we pre-train BERT \cite{devlin-etal-2019-bert} on a corpus that we collected in-house and a public Korean dialogue corpus\footnote{\url{https://aihub.or.kr/aihub-data/natural-language/about}}.
Our BERT consists of 12 layers, 768-dimensional embeddings, and 12 attention heads, resulting in total of 110M parameters.
The models except HyperCLOVA are based on Huggingface Transformers \cite{wolf-etal-2020-transformers}.
Naver Smart Machine Learning (NSML) platform~\cite{sung2017nsml, NSML} has been used in the experiments.

\subsection{Generator}
\label{sec:detail_generator}

For efficient training, we employ LoRA \cite{hu2021lora} for fine-tuning of all response generation and dialogue summarization models.
We fix adaptor rank to 4 and LoRA $\alpha$ to 32, with learning rate of $5 \times 10^{-4}$, weight decay factor of 0.1, and batch size of 8.
The maximum training epoch is 3 with early stopping. Training is completed within 10 hours using 1 NVIDIA A100.
The maximum sequence length is 2,048 and the inputs that exceed this length are truncated from the front.

\subsection{Retriever}
\label{sec:detail_ret}

Our retriever implementation is similar to context persona matching (CPM) method proposed in \citet{xu-etal-2022-long}.
The current dialogue context $D_t$ is encoded with dialogue encoder $E_D(\cdot)$, and each memory sentence $m_i \in M$ is encoded with memory sentence encoder $E_m(\cdot)$.
Here, $E(\cdot)$ refers to each input's representation, i.e. the encoder's output on the first input token ($[\mathrm{CLS}]$).
The encoders $E_D$ and $E_m$ are initialized with pre-trained BERT \cite{devlin-etal-2019-bert} architecture.
We use triplet loss to fine-tune the encoders as:
\begin{equation}
\max{(sim(D_t, m^+) - sim(D_t, m^-) + \alpha, 0)},
\end{equation}
where $m^+$ is a memory sentence matched with $D_t$ in the training dataset, $m^-$ is a memory sentence from other dialogue sessions in the training dataset, and $\alpha$ = 0.2 is the margin.
The models are trained for 20 epochs with early stopping using a maximum learning rate of $3 \times 10^{-5}$ and an linear scheduler.
This training takes about 3 hours using 1 NVIDIA V100.
At inference time, the top $k$ ($k =$ 5 in our experiments) memory sentences are retrieved from $M$ using cosine similarity:
\begin{equation}
sim(D_t, m_i)=cos(E_D(D_t), E_m(m_i)).
\end{equation}

\subsection{Memory Operator}
\label{sec:detail_operator}
For memory operation and NLI tasks, we define a unified text-to-text format of input and output to train our T5.
The input sequence becomes ``sentence 1: [$m$ or premise sentence] sentence 2: [$s$ or hypothesis sentence]'' and the target labels are mapped to single tokens corresponding to numeric characters $\{$``0'', ``1'', ``2'', ``3''$\}$; for memory operation, $\{$``0'': PASS, ``1'': APPEND, ``2'': REPLACE, ``3'': DELETE$\}$, and for NLI, $\{$``0'': Entailment, ``1'': Neutral, ``2'': Contradiction$\}$.
This makes it simple to transfer the model trained with NLI to the memory operation task by just replacing inputs and targets.

The models are trained with a batch size of 8.
They are trained for 20 epochs with early stopping, a maximum learning rate of $5 \times 10^{-5}$, and a linear scheduler.
This training takes about 1 hour using 1 NVIDIA A100.

\subsection{Hpyerparameter Search}
For all models, the learning rate was searched in the range of $[3 \times 10^{-5}, 5 \times 10^{-5}, 5 \times 10^{-4}]$ and the batch size in the range of $[8, 16, 32]$.
We tried at least 3 times for each setting to find the best configurations.

\section{Discussion}
\label{sec:discuss}
\subsection{Persona Updates in Our Dataset}
\label{sec:discuss_dataset}
Our dataset contains more persona updates, because of 1) relationship setting between interlocutors and 2) duration of episodes. 
 
According to \citet{altman1973social}, people disclose more private information as their relationship deepens. MSC \cite{xu-etal-2022-beyond} was collected in a setting of two strangers getting to know each other, where people tend to discuss rather easily shareable information such as their profiles or preferences. Our dataset reflects a more intimate relationship where people disclose more private realms of life, which is more likely to change over a few weeks (e.g. ``getting a physiotherapy'', ``gained a few pounds'', ``arguing with son'') or do not change frequently but will share updates if occured (e.g. ``got laid off'', ``got back together with partner'').

Also, the assumed duration of dialogue in our dataset is relatively longer than MSC \cite{xu-etal-2022-beyond}. The average term between sessions is about 2 days in MSC, while 10 days in our dataset. The entire span of each episode is 5 hours to 5 weeks in MSC, while 5 weeks to 10 weeks in our dataset. Thus, more persona information might change. 

\subsection{Memory Operation}
\label{sec:discuss_operation}
While collecting the dataset in Section \ref{sec:eval_update}, we found a few cases where the four proposed pairwise operations might not perfectly guarantee that the stored memory is lossless, consistent, and not redundant.
For example, when $m_i =$ ``Back hurts but hasn't seen a doctor yet'' and $s_j =$ ``Receiving physiotherapy at the hospital'', the resulting memory might not be lossless if only one of them is stored, while it might be inconsistent if both are left in memory.
In these cases, we ask the annotators to label them as {\footnotesize $``\text{FUSION}"$}, which might require combining the two sentences into a new sentence like ``Receiving physiotherapy at the hospital for back pain''. 
The cases of gold label = {\footnotesize $``\text{FUSION}"$} occur in about 1.09\% in the collected pairs, so we assume them to be negligible and discard them from our dataset in Table \ref{tab:pairwise_statistics}, since it would be costly to collect data exclusively for such cases and train a separate generator.
It might be helpful to use a generative approach that combines information from the two sentences into a new one $p(m'|m, s)$ selectively for such cases, though this would require additional consideration about the relationship between the newly generated sentence and other sentences.

Also, dependencies between memory sentences may exist.
For example, in the case of $m_i =$ ``got gastroenteritis'' and $m_j =$ ``The doctor banned alcohol, meat, and flour'', if the preceding sentence disappears, the following sentence should also be removed.
This case can be resolved simply by clarifying the fact that it is due to enteritis in the second sentence, but there may be some cases in which putting all the dependencies in a sentence is difficult.
A graph structure may be an alternative to design dependencies between memories.

\section{Variants of Response Generation Models}
\label{sec:eval_gen}

We compare various types of response generation models according to the training dataset and dialogue context type.
When using the original CareCall dataset, since this is a single session dataset, the model is always trained to predict $c_{t+1}$ given only $D_t$, i.e. $p(c_{t+1} | D_t)$.
When using the new CareCall$_{mem}$ dataset, the inputs vary depending on the context type.
\begin{itemize}
\item \textit{Without memory}: Given only $D_t$ to predict $c_{t+1}$.
\item \textit{Dialogue history grounded}: The dialogue histories from all previous sessions are concatenated before $D_t$.
\item \textit{Memory grounded}: The model described in Section \ref{sec:gen}, in which $M$ is concatenated before $D_t$. 
\end{itemize}

We evaluate the above models with 300 sessions of the human written dialogues described in Section \ref{sec:data_prep} as the test set (all $D_t$'s are second sessions of each episode).

Table \ref{tab:eval_gen} shows the results.
The model trained with CareCall dataset has no significant performance improvement in all metrics when the previous session history is given as a context.
Since CareCall dataset does not contain utterances that utilize memory, simply giving previous context at inference time does not enable the model converse using memory.
On the other hand, the newly collected CareCall$_{mem}$ dialogues are dependent on the previous session.
The model trained with such CareCall$_{mem}$ has a significant performance gain when the previous session history is given in the form of either raw dialogue history or summary.
Also, it is better to give a summary than to give a dialogue history, a finding consistent with a previous work \cite{xu-etal-2022-beyond}.
Furthermore, using CareCall + CareCall$_{mem}$ shows better or competitive results in all three types of session context.
This is probably because the general conversation performance is more advantageous with more data.
We also found that the utterances that explicitly mention memorized information are about 24.8\% of all utterances in the CareCall$_{mem}$ dataset, which is a reasonable proportion in real-life conversations.
This means that the general conversational ability is also important to predict responses in this task.
Therefore, we use CareCall + CareCall$_{mem}$ as the training dataset for response generation model on multi-session dialogues (Section \ref{sec:eval_multisession}).

\begin{table*}[t!]
\centering
\begin{adjustbox}{max width=\textwidth}
\begin{tabular}{lrrrr}
\toprule
\textbf{Model} & \textbf{PPL$\downarrow$} & \textbf{BLEU-1/2$\uparrow$} & \textbf{F1$\uparrow$} & \textbf{Dist-1/2$\uparrow$}\\
\midrule
\multicolumn{4}{l}{\textit{Without memory}} \\
CareCall & 13.214 & 0.155/0.054 & 0.214 & 0.074/0.178 \\
CareCall$_{mem}$ & 11.246 & 0.173/0.067 & 0.227 & 0.075/0.178 \\
CareCall + CareCall$_{mem}$ & 10.919 & 0.161/0.061 & 0.210 & 0.080/0.185 \\
\hline
\multicolumn{4}{l}{\textit{Dialogue history grounded}} \\
CareCall & 17.541 & 0.156/0.062 & 0.234 & 0.122/0.282\\
CareCall$_{mem}$ & 8.380 & 0.173/0.067 & 0.232 & 0.118/0.285 \\
CareCall + CareCall$_{mem}$ & 7.966 & 0.175/\textbf{0.079} & 0.238 & 0.107/0.244 \\
\hline
\multicolumn{4}{l}{\textit{Memory grounded}} \\
CareCall$_{mem}$ & \textbf{7.503} & 0.179/0.078 & 0.236 & 0.118/0.294 \\
CareCall + CareCall$_{mem}$ & 7.520 & \textbf{0.186}/0.075 & \textbf{0.239} & 0.110/0.279 \\
\bottomrule
\end{tabular}
\end{adjustbox}
\caption{Comparison of different context types and training dataset for response generation model.}
\label{tab:eval_gen}
\end{table*}

\section{Human Evaluation Metrics}
\label{sec:metric}
\begin{figure*}[t]
\centering
\includegraphics[width=\textwidth]{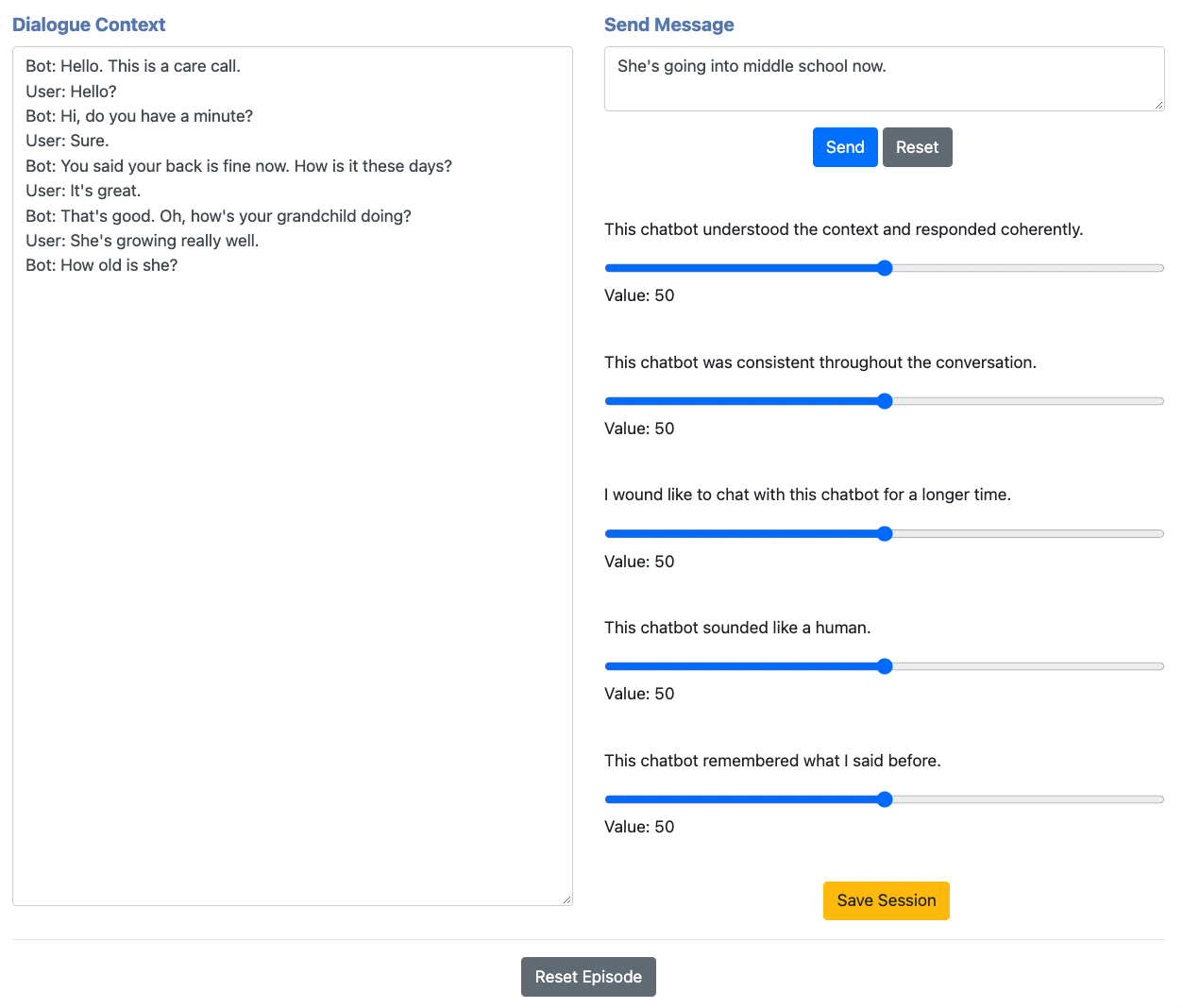}
\caption{Web-based user interface for the human evaluation. All texts are translated into English. For each session, crowdworkers can communicate with the system by sending messages. At the end of the session, they rate the degree to which they agree with the statements on a scale of 0-100 by moving the sliders. They converse with a randomly selected model for 5 sessions, and then reset episode to converse with another model.}
\label{fig:webui}
\end{figure*}

For each session of dialogue between human and a chatbot, we ask the crowdworkers to evaluate the quality of the chatbot by rating the degree to which they agree with the statement on each evaluation metric on a scale of 0-100 (inferface is given in Figure \ref{fig:webui}).
The statements for evaluation metrics are as follows.

\begin{itemize}
\item \textit{Coherence}: This chatbot understood the context and responded coherently.
\item \textit{Consistency}: This chatbot was consistent throughout the conversation.
\item \textit{Engagingness}: I wound like to chat with this chatbot for a longer time.
\item \textit{Humanness}: This chatbot sounded like a human.
\item \textit{Memorability}: This chatbot remembered what I said before.
\end{itemize}

The statements for the first four metrics are referred from previous literature \cite{li2019acute, finch-choi-2020-towards, ji-etal-2022-achieving, smith-etal-2022-human} for consistency of evaluation across different works. We additionally defined memorability to evaluate how well the model remembers previous conversations.

\begin{figure*}[t]
\centering
\includegraphics[height=0.95\textheight, width=\textwidth]{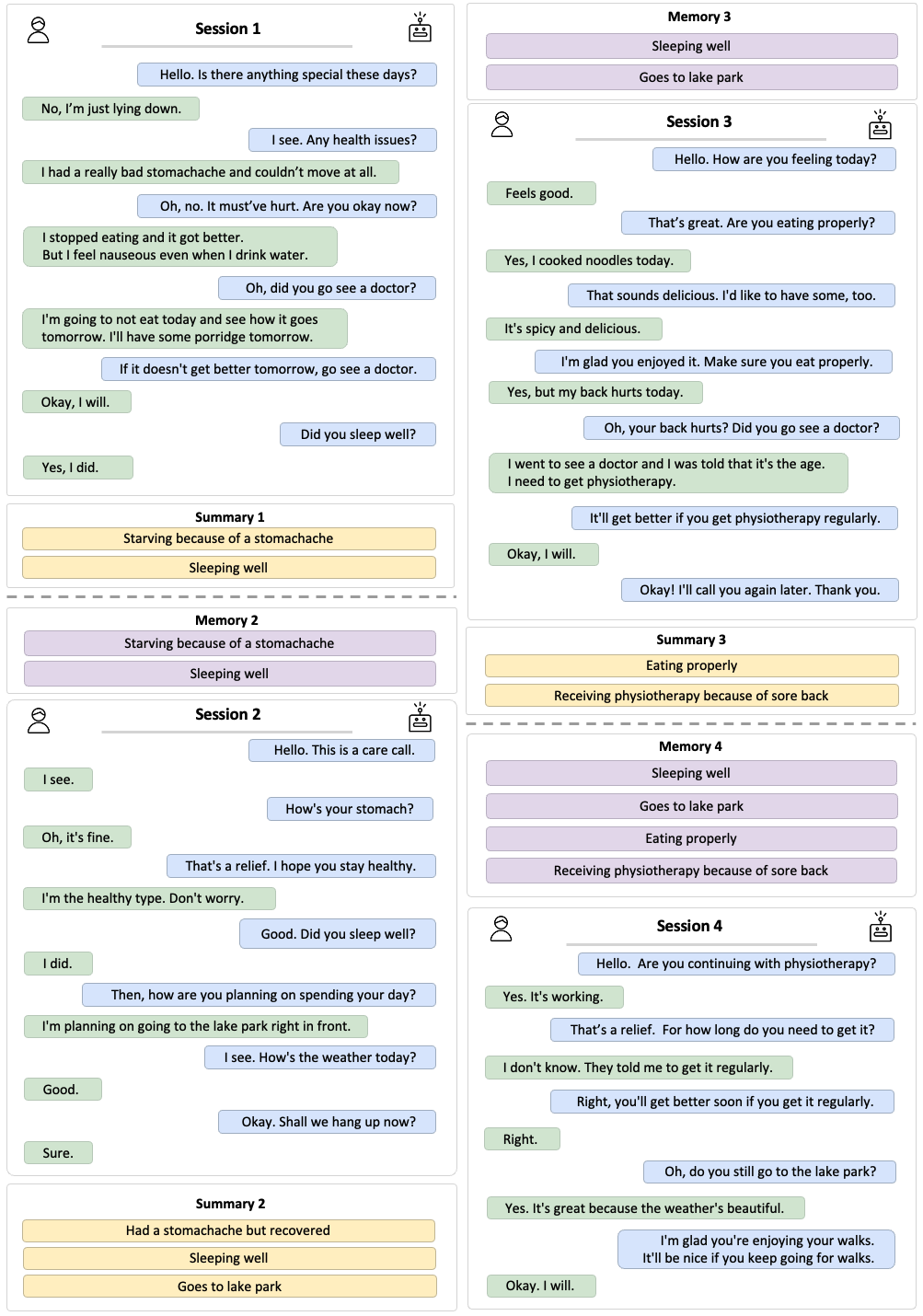}
\caption{Example four session conversation from the newly collected CareCall$_{mem}$ dataset. Memory 1 (an empty set) and Summary 4 are omitted. All the texts are translated into English.}
\label{fig:example}
\end{figure*}

\begin{table*}
\centering
\begin{adjustbox}{max width=0.8\textwidth}
\begin{tabular}{l|l|c}
\toprule
\textbf{Memory sentence $m$} & \textbf{Summary sentence $s$} & \textbf{label} \\
\midrule
Lost appetite and doesn't eat much & Lost appetite & PASS \\
\hline
Not sick & Doesn't have any particular health issues & PASS \\
\hline
Goes hiking every weekend & Goes hiking & PASS \\
\hline
Doesn't have any particular health issues & Had back surgery & REPLACE \\
\hline
Couldn't sleep well & Sleeping well after taking sleeping tablets & REPLACE \\
\hline
Living alone & Being with daughter for a while & REPLACE \\
\hline
Has a grandson in elementary school & Grandson enters middle school & REPLACE \\
\hline
Goes to the gym & Body is sore from exercise & APPEND \\
\hline
Eating properly & Receiving physiotherapy & APPEND \\
\hline
Has a dog & The dog likes carrots & APPEND \\
\hline
Has sleeping tablets prescribed & Has a son & APPEND \\
\hline
Gardening as a hobby & Many flowers are growing in the garden & APPEND \\
\hline
Had sore throat & Throat is fully recovered & DELETE \\
\hline
Takes pain relievers for a migraine & Migraine is gone & DELETE \\
\bottomrule
\end{tabular}
\end{adjustbox}
\caption{Example sentence pairs from the collected dataset in Table \ref{tab:pairwise_statistics}. All texts are translated into English.}
\label{tab:pairwise_example}
\end{table*}

\begin{table*}
\centering
\begin{adjustbox}{max width=\textwidth}
\begin{tabular}{l|l|rrrrr}
\toprule
\textbf{Session} & \textbf{Model} & \textbf{Coherence} & \textbf{Consistency} & \textbf{Engagingness} & \textbf{Humanness} & \textbf{Memorability} \\
\midrule
\multirow{3}{*}{Session 1} & Without memory & 0.1765 & 0.6926 & $-$0.0440 & 0.1058 & $-$0.2317 \\
 & Memory accumulate & 0.1765 & 0.6926 & $-$0.0440 & 0.1058 & $-$0.2317 \\
 & Memory update & 0.1765 & 0.6926 & $-$0.0440 & 0.1058 & $-$0.2317 \\
\midrule
\multirow{3}{*}{Session 2} & Without memory & 0.3553 & 0.6291 & $-$0.3753 & 0.1402 & $-$0.1257 \\
 & Memory accumulate & 0.5069 & 0.6565 & $-$0.1501 & 0.4418 & 0.3812 \\
 & Memory update & \textbf{0.6597} & \textbf{0.9033} & \textbf{$-$0.0724} & \textbf{0.5376} & \textbf{1.0263} \\
\midrule
\multirow{3}{*}{Session 3} & Without memory & $-$0.2383 & 0.4974 & $-$0.5080 & $-$0.3500 & $-$1.4041 \\
 & Memory accumulate & \textbf{0.2036} & \textbf{0.5614} & \textbf{$-$0.0640} & \textbf{0.2647} & \textbf{0.5017} \\
 & Memory update & 0.0854 & 0.5281 & $-$0.2186 & 0.0946 & 0.4338 \\
\midrule
\multirow{3}{*}{Session 4} & Without memory & $-$0.1135 & 0.4846 & $-$0.6073 & $-$0.3916 & $-$2.0036 \\
 & Memory accumulate & \textbf{0.2252} & \textbf{0.6787} & $-$0.5143 & $-$0.1339 & $-$0.2023 \\
 & Memory update & 0.1472 & 0.6541 & \textbf{$-$0.1663} & \textbf{0.1660} & \textbf{0.7231} \\
 \midrule
\multirow{3}{*}{Session 5} & Without memory & $-$0.3995 & \textbf{0.6622} & $-$0.7786 & $-$0.7298 & $-$1.6713 \\
 & Memory accumulate & $-$0.2753 & 0.4596 & $-$0.7616 & $-$0.3844 & $-$0.6674 \\
 & Memory update & \textbf{0.0598} & 0.2672 & \textbf{$-$0.3807} & \textbf{0.0599} & \textbf{0.2391} \\
\bottomrule
\end{tabular}
\end{adjustbox}
\caption{Per session average standardized scores of human evaluation metrics on live conversation. Since the models in the first session are all equivalent, scores are averaged for all models in Session 1.}
\label{tab:human_all}
\end{table*}

\begin{table*}
\centering
\begin{adjustbox}{max width=0.8\textwidth}
\begin{tabular}{l|rrrrr}
\toprule
\textbf{Model} & \textbf{Coherence} & \textbf{Consistency} & \textbf{Engagingness} & \textbf{Humanness} & \textbf{Memorability} \\
\midrule
Without memory & 71.79 & 94.52 & 54.67 & 81.37 & 36.37 \\
Memory accumulate & 73.86 & 96.10 & 54.98 & 82.68 & 50.31 \\
Memory update & 75.98 & 95.90 & 58.35 & 84.40 & 56.59 \\
\bottomrule
\end{tabular}
\end{adjustbox}
\caption{Average absolute scores of human evaluation metrics on live conversation for 5 sessions in each episode.}
\label{tab:human_abs}
\end{table*}

\end{document}